# Automated interictal epileptic spike detection from simple and noisy annotations in MEG data

Pauline Mouches[a], Julien Jung[a,b], Armand Demasson[a], Agnès Guinard[a], Romain Bouet[a], Rosalie Marchal[b], Romain Quentin[a]

[a] Université Claude Bernard Lyon 1, CNRS, INSERM, Centre de Recherche en Neurosciences de Lyon CRNL U1028 UMR5292, EDUWELL, F-69500, Bron, France.
[b] Department of Functional Neurology and Epileptology, Hospices Civils de Lyon.

**Corresponding author:**

Pauline Mouches
pauline.mouches@inserm.fr
CRNL - CH Le Vinatier - Bâtiment 452, 95 Bd Pinel, 69500 Bron, France
ORCID: 0000-0002-0304-5584

**Words:** 5363

**Abstract:**

In drug-resistant epilepsy, presurgical evaluation of epilepsy can be considered. Magnetoencephalography (MEG) has been shown to be an effective exam to inform the localization of the epileptogenic zone through the localization of interictal epileptic spikes. Manual detection of these pathological biomarkers remains a fastidious and error-prone task due to the high dimensionality of MEG recordings, and interrater agreement has been reported to be only moderate. Current automated methods are unsuitable for clinical practice, either requiring extensively annotated data or lacking robustness on non-typical data. In this work, we demonstrate that deep learning models can be used for detecting interictal spikes in MEG recordings, even when only temporal and single-expert annotations are available, which represents real-world clinical practice. We propose two model architectures: a feature-based artificial neural network (ANN) and a convolutional neural network (CNN), trained on a database of 59 patients, and evaluated against a state-of-the-art model to classify short time windows of signal. In addition, we employ an interactive machine learning strategy to iteratively improve our data annotation quality using intermediary model outputs. Both proposed models outperform the state-of-the-art model (F1-scores: CNN=0.46, ANN=0.44) when tested on 10 holdout test patients. The interactive machine learning strategy demonstrates that our models are robust to noisy annotations. Overall, results highlight the robustness of models with simple architectures when analyzing complex and imperfectly annotated data. Our method of interactive machine learning offers great potential for faster data annotation, while our models represent useful and efficient tools for automated interictal spikes detection.

**Keywords:** Magnetoencephalography, Epilepsy, Classification, Convolutional Neural Networks, Interactive machine learning,

# 1. INTRODUCTION

Epilepsy is one of the most commonly reported neurological diseases, with a prevalence estimated at around 50 million people worldwide (Bell et al., 2014). Uncontrolled epileptic seizures lead to reduced quality of life, increased risks of cognitive and psychiatric comorbidities, and a higher risk of premature death (Beghi, 2020). About one-third of epilepsy cases remain drug-resistant. For these cases, surgical intervention focused on the brain regions triggering epileptic seizures (i.e. epileptogenic zone) can be considered (Ryvlin et al., 2014). In practice, the localization of the epileptogenic zone is inferred from several sources of information including symptom observations and several imaging medical exams. During the pre-surgical evaluation, the use of magnetoencephalography (MEG) was shown to improve diagnosis and treatment outcomes (Rampp et al., 2019). MEG non-invasively records neural activity with a high temporal and spatial resolution. MEG captures interictal epileptic spikes, which are short pathological events, of about 80ms, that are well-established biomarkers of epilepsy (Jung et al., 2013). These spikes propagate within a network centered on the epileptogenic zone. Therefore, accurately detecting spikes in MEG allows the epileptogenic zone to be located (Jung et al., 2013). Interictal spike detection in MEG is most often performed manually by analyzing several minutes of recordings over several hundreds of MEG sensors (De Tiège et al., 2017). However, the high dimensionality of MEG data, along with the variability across clinicians, patients, and epilepsy types, can result in subjective and inaccurate interictal spike annotations. While no study exhaustively assessed interrater agreement on MEG spike annotations, the difficulty of the task can be compared to interictal spike detection in EEG, for which (Jing et al., 2020) reported only a fair interrater agreement (Gwet's κ = 48.7) among 8 experts when analyzing individual spike events.

Automated methods for interictal spike detection have been proposed, but they predominantly focus on scalp electroencephalography (EEG) data, which is more commonly acquired in patients with epilepsy (da Silva Lourenço et al., 2021; He et al., 2021; Mohammed et al., 2023). Such methods can, in principle, be transposed to MEG data, although the number of sensors is typically higher in MEG (16-200 in EEG, 200-300 in MEG), which increases data complexity. Moreover, substantial differences in spike duration and morphology between MEG and EEG have been reported (Ossenblok et al., 2007). Machine learning-based methods offer several advantages as they require less a priori knowledge about the characteristics of interictal spikes, which can be automatically optimized using deep learning (Abd El-Samie et al., 2018). Recent deep learning studies have focused on tasks related to either EEG data or seizure detection, for which data are more commonly available. To date, only few deep learning models specifically designed for interictal spike detection in MEG have been proposed (Fernández-Martín et al., 2024; Hirano et al., 2022; Zheng et al., 2020). All studies used convolutional neural networks (CNNs) and reported theoretically high performance. However, their limitations include either a small sample size (10 to 35 patients) (Fernández-Martín et al., 2024; Zheng et al., 2020, 2023) or an exclusive focus on typical spikes, which are easier to detect, thereby limiting the applicability to clinical settings (Fernández-Martín et al., 2024; Hirano et al., 2022). Moreover, most methods require channel-wise annotation of the data, whereas clinicians typically only report the timing of the spikes. This last and major constraint also applies to most feature-based machine learning methods (Alotaiby et al., 2017). Finally, existing methods demonstrated poor performances when tested on heterogeneous real-life data in our previous work (Mouches et al., 2024). Indeed, directly comparing results across studies without model re-training remains challenging, as datasets, sample sizes, and definitions of what qualifies as a spike vary significantly between studies (da Silva Lourenço et al., 2021). In summary, despite these advancements, an accurate interictal spike detection algorithm suitable for clinical use is still lacking.

Interactive machine learning is a human-in-the-loop strategy that incorporates human feedback during the model training process. This approach enables back-and-forth exchanges between the model and the experts to refine the model iteratively, and potentially adapt it more closely to a specific expert's preferences (Amershi et al., 2014). In the medical domain, annotation is known to be error-prone and costly to acquire. Therefore, evaluating the benefits of interactive machine learning in this domain is highly relevant as it offers a valuable opportunity to improve data annotation quality while requiring limited expert time (Göndöcs & Dörfler, 2024; Mosqueira-Rey et al., 2023). Interactive machine learning, and more specifically interactive refinement of model output, has been explored mainly for medical image segmentation tasks, in which refinement is often done through expert scribbles or seed points. However, much less development has been devoted to regression and classification tasks (Budd et al., 2021). In the context of EEG and MEG, the use of interactive machine learning is mainly limited to brain-computer interface applications (Mosqueira-Rey et al., 2023). One study implemented this procedure in a classification context for EEG artefact detection (Diachenko et al., 2022) and showed mitigated results, mostly demonstrating increased model confidence when using expert revised data. While interactive learning approaches are often expected to enhance both annotation quality and model performance, it remains unclear whether such expert refinement provides measurable benefits when models are robustly trained on noisy annotations.

In this study, we propose two machine learning-based methods for interictal spike detection from MEG data. The first method is a feature-based artificial neural network (ANN) that uses simple, well-known signal-based features, and the second is a lightweight convolutional neural network (CNN) trained directly on raw data, as illustrated in Figure 1a). We evaluate our methods while employing an interactive machine learning strategy to improve MEG data annotation quality. To this end, we employed an imperfectly annotated database of 72 patients to iteratively train models and retrieve expert feedback on the model outputs. Through this iterative process, expert annotators refined the annotations across our entire database allowing us to evaluate the importance of annotation quality on a holdout test database with high-quality annotations. This work represents the first attempt to train and test machine learning models from routinely acquired MEG data, with the aim of detecting interictal epileptiform discharges, using only temporal, single-expert, and non-exhaustive clinical annotations. This approach eliminates the need for channel-wise labeling and enables learning directly from real-world clinical data, making it more relevant for future clinical application.

## 2. METHODS

### 2.1. Data

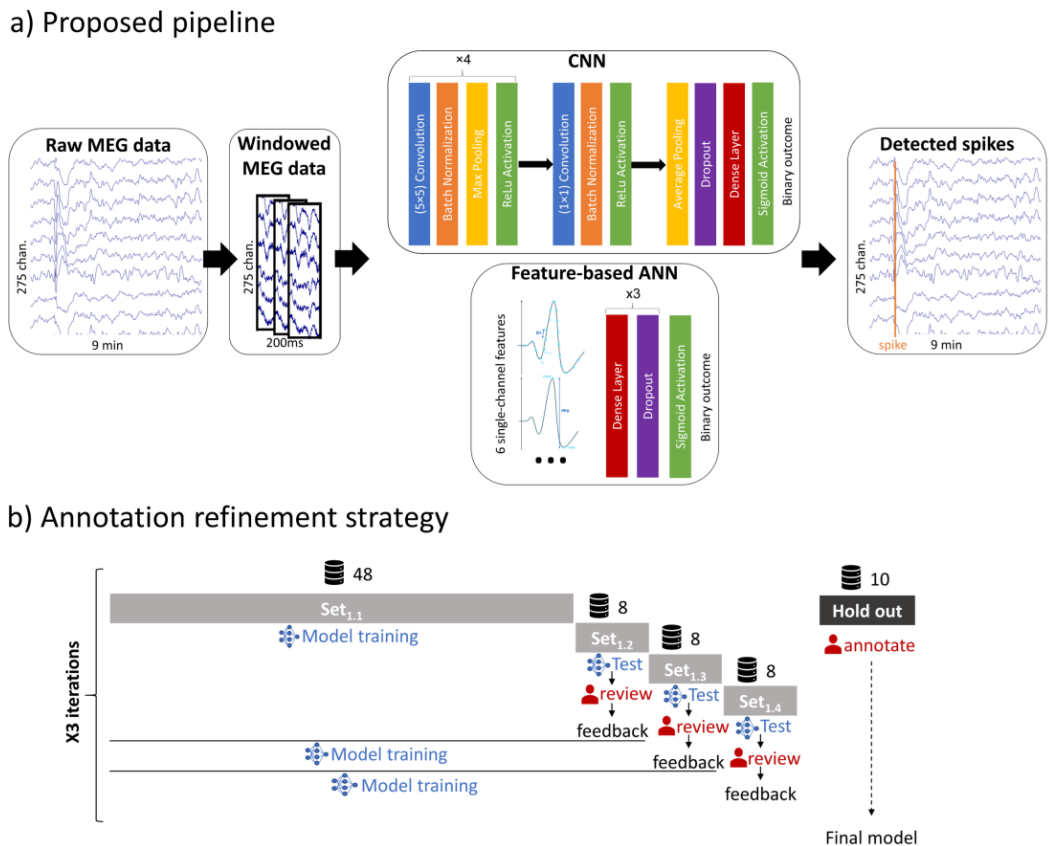

*Fig. 1: a) Graphical illustration of the proposed pipeline for automated interictal spike detection. b) Graphical illustration of the annotation refinement procedure. For each set, the number of patients is specified above the box and the purpose of the set is specified below. Three iterations are performed so that all patient data is reviewed once.*

The database contains MEG recordings of 82 drug-resistant epilepsy patients, aged between 5 and 65 years old (mean: 22 years old), acquired as part of their presurgical evaluation at the Hospice Civil de Lyon (HCL). All patients gave informed consent and the HCL approved the study on October 2012 (study NCT01735032: Multimodal Imaging in Pre-surgical Evaluation of Epilepsy (EPIMAGE)). 45-minute resting-state recordings with eyes closed were acquired using a CTF Omega 275-channel whole-head system (VSM MedTech Ltd.) at a sampling frequency of 1200 Hz. The recordings were saved in 15 files of 3 minutes each.

Interictal spikes were initially annotated using the DataEditor software (part of the CTF software bundle), which uses the Synthetic Aperture Magnetometry with excess kurtosis (SAMg2) technique (Scott et al., 2016). The main goal was to identify each patient's prototypical spikes. For most patients, the annotations focused on capturing a sufficient number of high-amplitude, reproducible spikes to enable accurate source localization. Consequently, the selection was not exhaustive and primarily aimed at reducing the false-positive rate. In some cases, datasets included only a few spikes of uncertain pathological relevance that could be considered as false-positives. These spikes were retained in the initial annotation when their presumed localization aligned with findings from other investigations (e.g., MRI brain lesions, focal hypometabolism on 18FDG PET, or clinical semiology). The reannotation process is intended to correct this type of error, which can be hard to spot initially.

### 2.2. Preprocessing

Among the 45-minute recordings split into 3-minute recording files, only the three 3-minute recording files with most spikes were kept, based on these initial annotations, for each patient. This reduction of the data quantity aimed to accelerate the training process of our models and the reannotation procedure, while preserving the richness of the data in terms of inter patient variability. Minimal signal preprocessing was applied in order to obtain models that are robust to diverse artefacts. Selected datasets were bandpass filtered (0.5-50Hz) and resampled to 150Hz, following usual processing steps (Roy et al., 2019). 200ms sliding windows were then cropped with a 60 ms overlap on each side. This duration was chosen regarding the full duration of an interictal spike ranging from 70 ms to 100 ms. Windows containing a spike annotated in the window center (30ms away from the borders) were labelled as "*spike*" windows while the others were labeled as "*spike-free*" windows. Doing so ensures that most of the spike duration is included in the windows labelled as containing a spike, while each spike is included in a single window and not replicated in the database. Three expert annotators (JJ, RB, RM) trained for interictal spike detection by the same expert neurologist, participated in the interactive machine learning procedure. Conversely, the holdout test database was reannotated from scratch by a single expert neurologist (JJ).

### 2.3. Problem definition and models

Each window can be denoted as $X_i \in R^{(n_s, n_t)}$ where $n_s$ corresponds to the number of MEG sensors and $n_t$ to the number of time points. Sample labels are denoted as:

$$y_i = \begin{cases} 0 \text{ for } \textit{spike-free} \text{ windows} \\ 1 \text{ for } \textit{spike} \text{ windows} \end{cases} \quad (1)$$

Two-dimensional convolutional neural network (2D CNN): We used a lightweight simple two-dimensional CNN, inspired from our previous work (Mouches et al., 2024), which is the most common model type used for EEG signal analysis (Roy et al., 2019). Our choice was suggested by previous studies demonstrating that simple lightweight models tend to overperform more complex models when used on a highly heterogeneous database (El Ouahidi et al., 2024). Our CNN contains 5 convolutional blocks made of a convolutional layer with a (5x5) kernel followed by a batch normalization layer, a (2x2) max pooling layer and leaky ReLu activation. The convolutional layers from the blocks have the following number of filters (32, 64, 128, 256, 64) and kernel L2 regularization of $10^{-3}$ was used to avoid overfitting. Finally, we applied two-dimensional average pooling, dropout (rate = 0.5), and a dense layer with sigmoid activation (see Figure 1 a).

Feature-based artificial neural network (ANN): We propose an artificial neural network (ANN) to evaluate classical handcrafted time-domain features as an interpretable baseline for MEG spike detection. Such feature-based approaches have been widely used in spike detection. Following this line of work, 6 single-channel hand-crafted time-domain features similar to those typically reported in previous studies (Abd El-Samie et al., 2018), were extracted. Over each sensor $S$, for each time window ranging from time 0 to time $T$, we computed:

- The maximum peak to peak amplitude:

$$\max_{t \in \{0,\dots,T\}}(S_t) - \min_{t \in \{0,\dots,T\}}(S_t) \quad (2)$$

- The maximum upslope:

$$\max_{t \in \{0,\dots,T-1\}}(S_{t+1} - S_t) \quad (3)$$

- The maximum downslope:

$$\min_{t \in \{0,\dots,T-1\}} (S_{t+1} - S_t) \tag{4}$$

- The average slope, as defined in (Tarassenko et al., 1998):

$$\max_{t \in \{0,\dots,T-2\}} \left( \frac{|S_{t+1} - S_t| + |S_{t+2} - S_{t+1}|}{2} \right) \tag{5}$$

- The sharpness:

$$\max_{t \in \{0,\dots,T-2\}} |(S_{t+2} - S_{t+1}) - (S_{t+1} - S_t)| \tag{6}$$

- The standard deviation over time.

These features were concatenated across sensors and fed to a fully connected ANN made of three dense layers (512, 128, 16 neurons), with ReLu activation, each followed by dropout (rate = 0.2). A last dense layer with 1 neuron and sigmoid activation was used to obtain the binary prediction (see Figure 1 a)).

State-of-the-art model: We re-implemented the EMS-Net model proposed by Zheng et al (Zheng et al., 2020) and trained it on our data. Briefly described, this model extracts both local (single sensor) and global (multi sensor) features, using convolutional layers. EMS-Net was initially designed to work on subgroups of sensors according to their cortical regions. However, as our annotations are not region specific, we used it considering all sensors together as a unique subgroup. Although this configuration slightly deviates from the originally proposed architecture, EMS-Net remains the main deep learning-based state-of-the-art-model from the literature, and thus, a fair comparison.

Optimization: All three models were optimized using the Adam optimizer with a learning rate of $10^{-4}$, trained with a batch size of 32 and a class-weighted binary focal cross-entropy loss function (Lin et al., 2017) to overcome the very high imbalance ratio of our data. Class weights were computed as follows: $W_i = (1/N_i) \times (N/2)$, with $N$ representing the sample size for the class of interest $i$. Furthermore, the focal loss (FL) allows the model to focus on samples that are harder to classify:

$$FL(p) = \begin{cases} -\alpha(1-p)^\gamma \log(p) \; if \; y = 1 \\ -(1-\alpha)p^\gamma \log(1-p) \; if \; y = 0 \end{cases} \tag{7}$$

where $p$ corresponds to the probability output of the last sigmoid dense layer of the model. $\alpha$ was set to 0.25 and $\gamma$ to 2, based on empirical tests. Optimal number of epochs were determined using early stopping on the validation loss, using 15% of the windows as validation data without stratifying per patient due to the limited number of subjects and the need to ensure sufficient samples from each spike morphology for model optimization.

## 2.4. Interactive machine learning for annotation refinement

Among the initial database of 82 patients, 10 patients with good signal quality were selected as holdout test set. Their recordings were annotated from scratch by the clinicians and were not used in any step of model training. The other 72 patients were used for the annotation refinement procedure illustrated on Figure 1 b), using the 2D CNN model architecture. The refinement procedure is carried out over 3 iterations. For each iteration $i$, the database was split into 4 subsets. The first set ($Set_{i,1}$) was made of 2/3 of the data (48 patients), while $Set_{i,2}$, $Set_{i,3}$, and $Set_{i,4}$ contained 1/9 of the data each (8 patients). Within that iteration, a first model was trained on $Set_{i,1}$ and tested on $Set_{i,2}$. The test results were sent to the experts for review. The review consisted of three possible actions: validating or not a detected spike, validating or not a spike that was annotated in the original annotations, and adding a new spike that was neither detected by the model nor contained in the original annotations. After receiving the expert feedback on $Set_{i,2}$, a second model was trained using data from both $Set_{i,1}$ and $Set_{i,2}$, and was tested on $Set_{i,3}$. Expert feedback was then collected on $Set_{i,3}$, followed by a new model training step on data from $Set_{i,1}$, $Set_{i,2}$, and $Set_{i,3}$. This last model from the current iteration was tested on $Set_{i,4}$ and expert feedback was collected, resulting in 3/9 of the data (24 patients) being reannotated by the experts per iteration. In the following sections, all models trained during the annotation refinement procedure are called "weak models".

**2.5. Model architectures comparison**

The three model architectures were trained at the end of the different iterations of the reannotation procedure. After each iteration, all data were used for training considering the refined annotations for already reannotated patients and the original annotations for the others. Their performances were then compared when tested on the holdout test patients.

**2.6. Impact of annotation quality on model training**

In order to better understand the impact of annotation quality on the proposed spike detection model performance, we designed a second experiment consisting of progressively adding random label noise into the training data, and testing the model on the 10 holdout test patients. Specifically, for each noise level, a predefined proportion of windows was selected for label inversion. The selection was performed after pooling all windows from all patients, without stratification at the patient level. The original class ratio of *spike* and *spike-free* windows and the total number of windows were preserved. This experiment was repeated three times with different seeds to ensure robustness of the results.

**2.7. Evaluation**

Model performances are reported using a relaxed version of the specificity, sensitivity and F1-score, which corresponds to the harmonic mean between precision and recall, for the spike class. The relaxed version of these metrics authorizes a one window shift of the detection compared to the ground truth. Therefore, if a spike is annotated on the window $X_t$, but detected on the $X_{t-1}$ or $X_{t+1}$ window, it is considered as correctly detected (Gao et al., 2021). Such case is realistic when the spike spans across windows or when the expert annotation is not exactly centered on the spike.

For the 2D CNN model, explainability analysis using saliency maps were conducted. Saliency maps show the areas of the signal used by the model to make its prediction and can be used to visually assess the relevance of the model. The SmoothGrad algorithm (Smilkov et al., 2017) was used. This method consists in applying random noise from a normal distribution N(0,0.1) to each test sample, corresponding, in our case, to windows, before passing it through the trained model. The partial derivative of the loss between the model prediction and expected output is backpropagated through the model, resulting in one gradient value per signal time point for each sensor, representing importance. This process is repeated 10 times for each window and the output saliency maps are averaged.

## 3. RESULTS

**3.1. Annotation refinement**

<u>Spike labels:</u> The annotation refinement procedure led to a considerable increase in the number of spikes annotated in the data going from 3323 (average of 48.9 spike per patient) in the original annotations to 4664 (average of 68.6 spike per patient) in the refined annotations. Among the original annotations made by the experts, 2094 spikes were still considered correct after receiving expert feedback, while 1229 (~37%) were considered incorrect, experts revising their previous annotations in those cases. During the annotation refinement procedure, four patients were excluded due to excessive noise in their MEG recordings. After the reannotation procedure, 9 patients ended up having less than 10 spikes and were therefore not used in further experiments comparing model architectures.

Figure 2 illustrates the main results of the annotation refinement procedure while showing the differences between the three experts who participated. Overall, we observed a high variability in the number of spikes detected between patients and the third expert (RB) demonstrates higher rate of validated spikes compared to the two others (Figure 2a). We also observed that the number of detected (Figure 2b) and validated spike (Figure 2c) correlates with the number of spikes in the original annotations (magenta triangles), indicating that original annotations quality was reasonable.

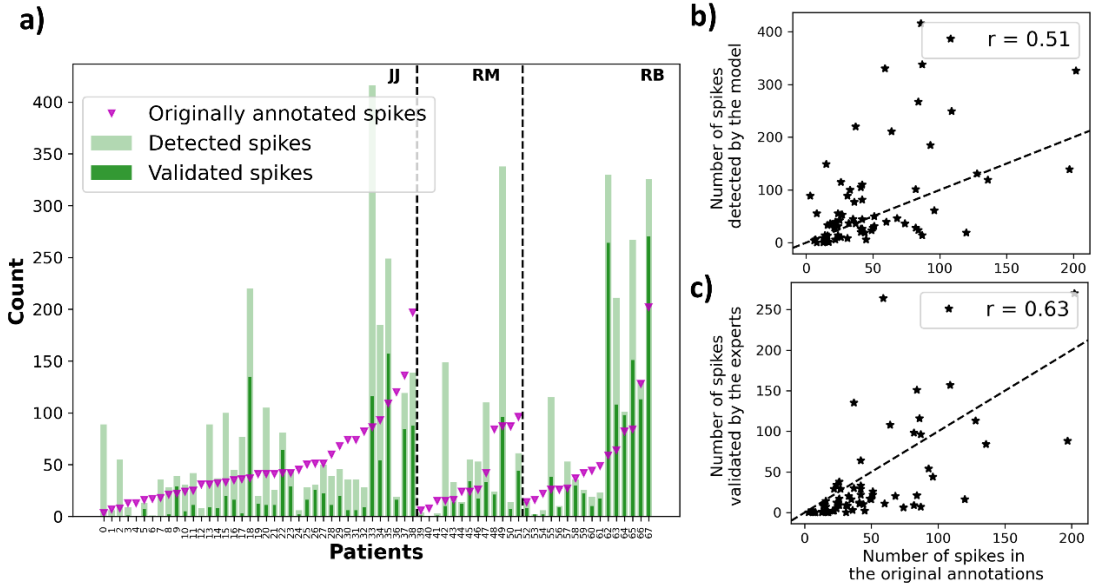

*Fig. 2*: a) Histogram representing the number of spikes present in the original annotations, detected by the weak models during the reannotation procedure, and further validated by the expert during the review, for each patient (9 min recordings). Vertical dashed lines delimitate patients that have been reviewed by the three different experts. b), c) plots illustrating the number of spikes detected by the weak models (b) and validated by the experts (c) versus the number of spikes in the original annotations. Each dot represents a patient.

<u>Weak model spike detection:</u> To further evaluate the utility of the 2D-CNN-based weak models in the interactive machine learning process, we assessed their respective performance at the time each model was inferred for review on the test patients included in the corresponding iteration of the annotation refinement procedure. This setup allows us to compare model predictions against both the original annotations and the refined annotations for each patient involved in the annotation refinement procedure (see Table 1).

|  | F1-score | Sensitivity | Specificity |
|---|---|---|---|
| **Original annotations** | 0.23 ±0.18 | 0.29 ±0.26 | 0.99 ±0.018 |
| **Refined annotations** | 0.42 ±0.27 | 0.49 ±0.30 | 0.99 ±0.012 |

*Table 1*: Spike detection performance across patients when compared to the original or refined annotation. Values are reported as mean ± standard deviation across patients.

Results showed a substantial increase in performance when comparing the model detection to the refined annotations, with the F1-score almost doubling (from 0.23 to 0.42). This improvement could reflect the model's ability to robustly learn spike patterns, allowing it to identify spikes that were not originally annotated, but it may also be influenced by the refinement process itself: the experts could have been guided by the model detections, potentially validating false spikes. At the same time, the fact that they also added many spikes neither detected by the model nor present in the original annotations demonstrates their careful attention. We also observed a high inter-patient variability with a large standard deviation across patients for all metrics. Figure 3 further illustrates the relaxed F1-score metric of weak model predictions against the original and refined annotations, for each patient. One patient with no spike detected by the model and no annotated spike was given a F1-score of 1. Most of the patients exhibited higher relaxed F1-score when model detections were compared to the refined annotations (green), meaning that the expert validated a substantial amount of detection that were not originally present in the annotations. Only 3 patients obtained a lower F1-score when evaluated against the refined annotations.

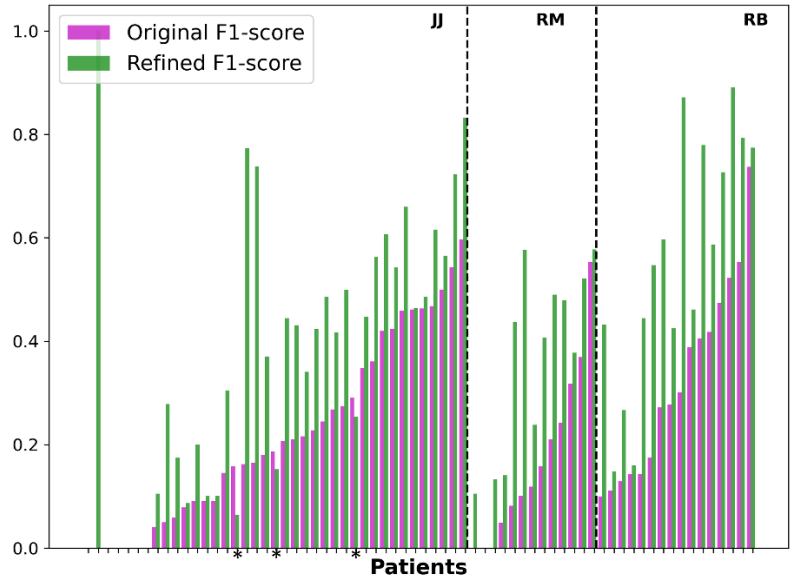

*Fig. 3*: *Comparison of weak model F1-scores for each patient when evaluated against original and refined annotations. Vertical dashed lines delimitate patients that have been reviewed by the 3 different experts. Stars indicate patients with a lower F1-score when evaluated against the refined annotation.*

**3.2. Model architectures comparison on the holdout test patients**

The holdout data are representative of the rest of the data with 658 spikes annotated by the expert (average of 65.8 per patient). Models were trained using recordings from 59 patients, after the exclusions of 13 patients as detailed in section 3.1. Figure 4 illustrates the model performances at each iteration. Iteration 0 corresponds to none of the Set being reannotated yet, while iteration 3 corresponds to annotations from all Sets being refined. The experiment was repeated 3 times with different initializations seeds for the model weights and training/validation data split, and the mean and standard deviation across repetitions is reported.

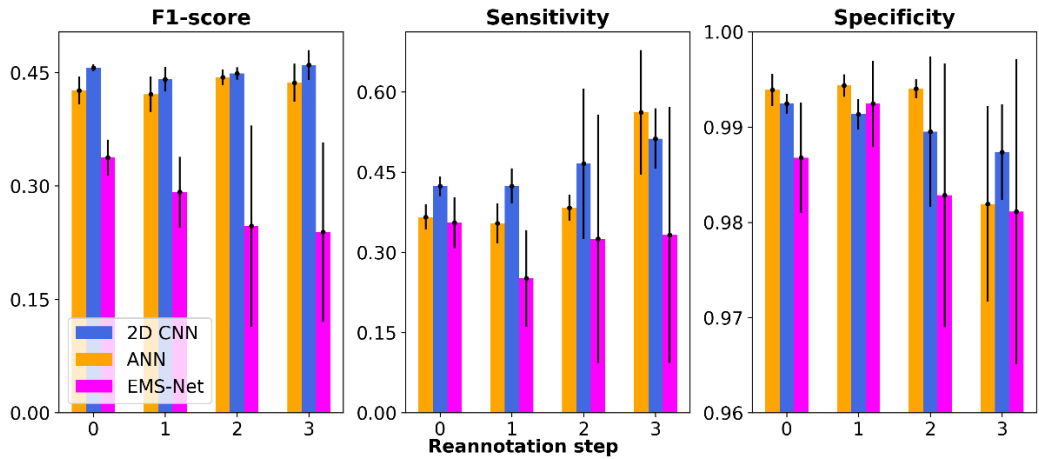

*Fig. 4*: *Model results on the holdout test data after each iteration of the annotation refinement procedure, as illustrated on Figure 1 b). Iteration 0 corresponds to the case where none of the patient data were reannotated yet. Mean and standard deviation across 3 repetitions with different seeds, for each reannotation step, are reported.*

We first observed that the proposed 2D CNN and feature-based ANN perform very similarly (best relaxed F1-scores CNN: 0.46; ANN: 0.44), and much better than the EMS-Net model. The relatively low performance of the EMS-Net model may be explained by the fact that its architecture was originally designed to work on groups of sensors from a specific brain region, which were annotated separately. In our case, our annotations do not contain such level of detail and simply describe the timing of spikes and not their localization. Across the annotation refinement steps, model performances remained stable, except for the EMS-Net model, which showed a decrease in F1-score and an increased variability across repetitions. For both our models, we observed a slight improving trend in sensitivity, indicating that more spikes are detected.

Examples of recordings with spikes detected by the two best-performing models (2D CNN and feature-based ANN) are shown in Figure 5, highlighting the models' high specificity, as large portions of the recordings contain no detected spikes. It also illustrates the difficulty of the annotation task when spikes are very close to each other's, thus justifying the need for relaxed metrics. Finally, we observed that, although both models obtain similar performances, they do not necessarily detect the same spikes.

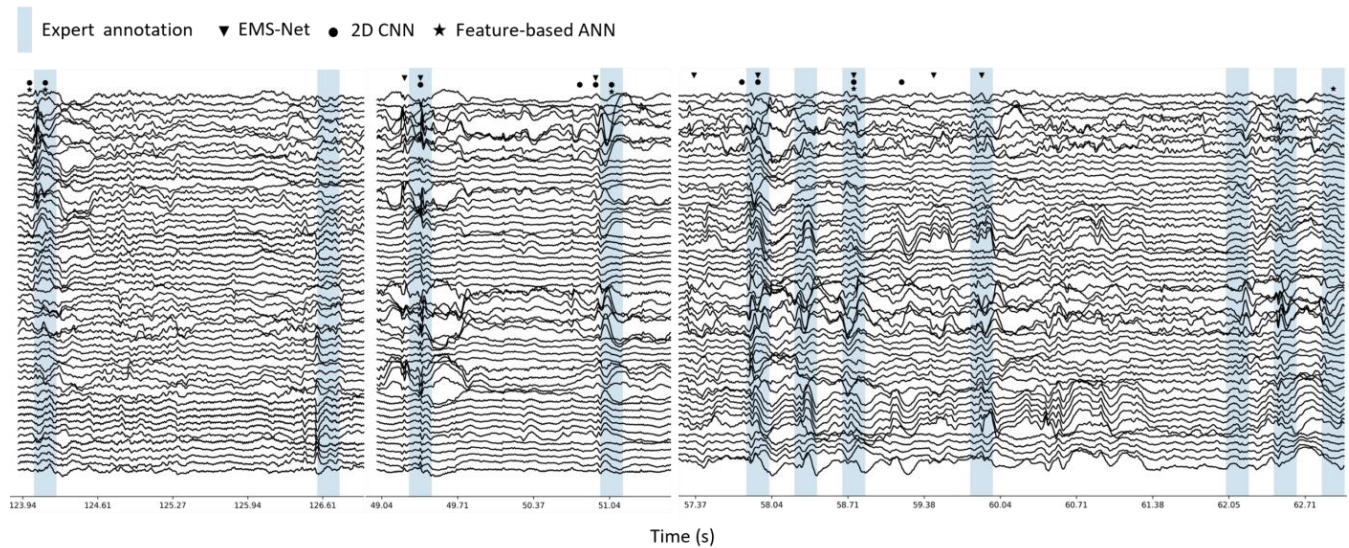

**Fig. 5**: *Example of detected spikes for 3 patients. One sensor out of five is shown. Expert-annotated 200 ms spike segments are highlighted in blue, while markers above the signal indicate the centers of 200 ms segments detected as containing spikes by the different models.*

**3.3. Impact of annotation quality on model training**

The results on Figure 4 suggested that our annotation refinement procedure did not improve our model's performance on the holdout testing set after retraining with the refined annotations. To further investigate the impact of imperfect annotations on model training, we conducted a random noise addition experiment using the two best performing models, described in section 2.6. As the class distribution between *spike* and *spike-free* windows is highly imbalanced, the percentage of noise corresponds to the percentage of samples that were corrupted regarding the total number of spikes (e.g., at 100% noise level, all N *spike* window labels are flipped, along with an equal number N of randomly selected *spike-free* windows). For both models, we observed a sharp drop in performance for the random noise experiment, after 10% of corrupted annotations (see Figure 6). High standard deviation across repetitions with different seeds is also observed in results with high noise level.

These experiments also allow us to assess the quality of our original annotations (see section 2.1). After being reviewed by experts, 37% of these original annotations were considered as incorrect (see section 3.1). Thus, direct comparison can be made between the performance of the model trained on the original annotations (Figure 4, reannotation step 0) and the performance of the model trained with 37% of corrupted annotations (Figure 6). The lower performance in the model with 37% corrupted annotations suggests that errors in the original annotations were not random but often resembled true spikes.

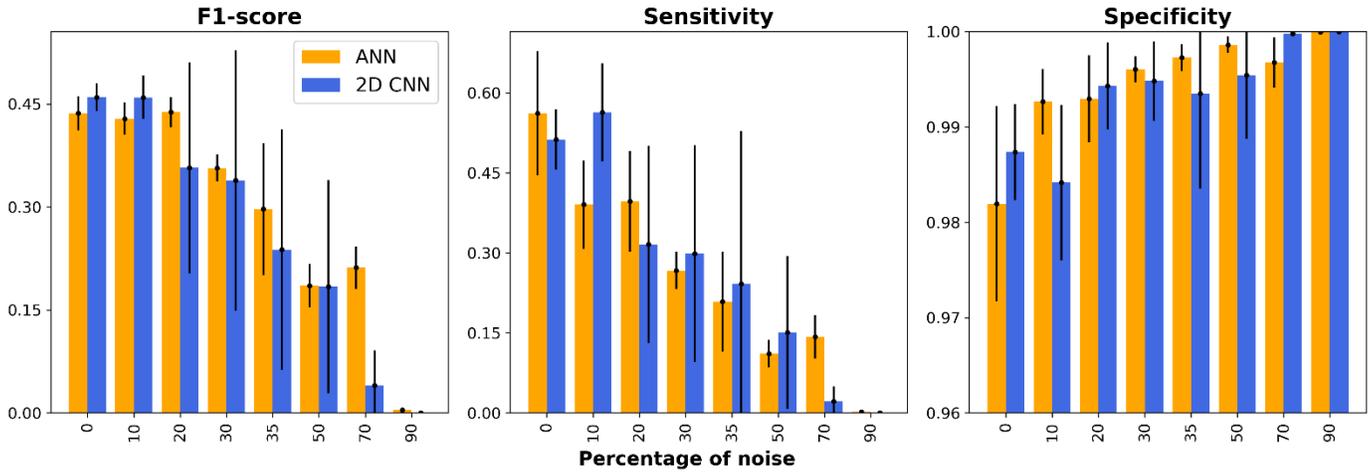

*Fig. 6: Results of the random label noise addition experiment. Annotations of a given percentage of random windows were corrupted, models were then retrained and tested on the holdout test data. For each noise level, mean and standard deviation across three repetitions with different seeds are reported. A noise level of 37% corresponds to corrupting the same proportion of data as the amount of windows that changed labels between the original and refined annotations.*

### 3.4. Saliency maps

Explainability analysis of the 2D CNN reveals which parts of the signal contributed most to the predictions made by the model. Figure 7 illustrates examples of spikes that were correctly detected (true positives) and spikes that were incorrectly detected (false positives), for one of the test patient. Model probability estimates can be interpreted as a score of model confidence. The highest gradient values, representing importance, are seen around peaks of the signal that appears on adjacent channels, which matches with clinical knowledge about interictal spikes. Visualization of false positives highlights the difficulty of the task as the model focuses on relevant signal features that highly look like interictal spike patterns. Overall, this visualization ensures us that the model learned correct features typically attributed to interictal spikes in clinical knowledge.

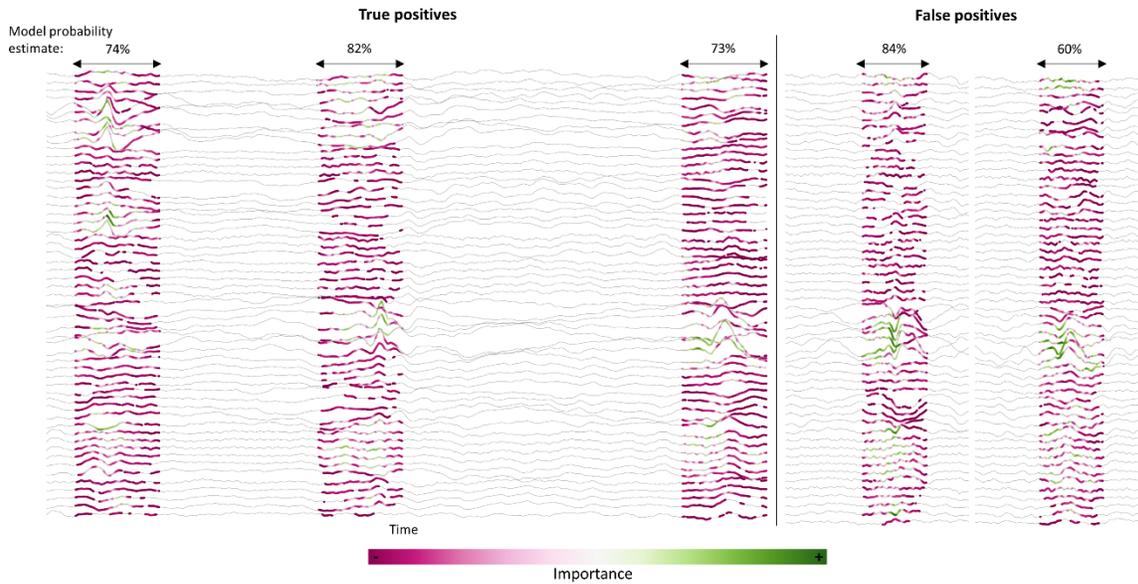

*Fig. 7: Saliency maps. The left panel shows importance values for spikes that were correctly detected, while the right panel shows false positives. One sensor out of five is shown.*

## 4. DISCUSSION

In this study, we proposed two machine learning models for interictal spike detection and evaluated them using an interactive machine learning strategy to improve the quality of MEG data annotations. Main results regarding model performances demonstrate that both our feature-based ANN and CNN models performed similarly and outperformed the state-of-the-art method. These findings support the idea that simple models may generalize better and handle the high variability of the data. Our main finding regarding the interactive machine learning strategy is that both models are robust to noisy annotations, as their performance does not significantly improve when trained with higher-quality annotations. This robustness presents promising opportunities for efficiently retrieving more non-exhaustively annotated data to train future models.

Machine learning for epileptic spike detection: Our CNN achieved the highest performance (relaxed F1-score during the last reannotation step = 0.46, Figure 4), while the feature-based ANN also performed well (relaxed F1-score during the last reannotation step = 0.44, Figure 4), suggesting that the selected features effectively capture most of the detectable spikes. Our model yields performance values lower than those reported in some previous studies(Fernández-Martín et al., 2024; Hirano et al., 2022; Zheng et al., 2020). For instance, the EMS-Net study (Zheng et al., 2020) reported an F1-score of 93% for individual interictal spike detection. However, expert inter-rater agreement for individual spike detection in EEG is known to be only fair (Gwet's κ = 48.7) (Jing et al., 2020). Although such agreement has not been formally assessed for MEG, it is likely of similar or even lower magnitude given the higher dimensionality of MEG signals (hundreds of channels versus a few dozen in EEG). Therefore, F1-scores from deep learning models that substantially exceed known inter-rater agreement levels should be interpreted with caution regarding their clinical relevance (Kljajic et al., 2025). When trained and tested on our data the performance of the EMS-Net model is lower than the performance of our models (relaxed F1-score during the last reannotation step = 0.24, Figure 4). This finding is in line with our previous work, where we showed that simple models tend to perform best for such epileptic spike classification when applied to real life data (Mouches et al., 2024). A similar observation was recently reported for EEG data in brain computer interface applications where a simple deep learning model outperformed more complex architectures, especially in cross-subjects experiments, where data heterogeneity between subjects is high (El Ouahidi et al., 2024). Overall, such finding is commonly observed in the medical field (Peng et al., 2021; Raghu et al., 2019).

Interactive machine learning for brain neural recordings classification: Our proposed interactive machine learning procedure provides a valuable opportunity to accelerate data annotation, especially when expert time is limited. Manual review requires about 4 minutes of expert time per minute of recording, as each ~ 500 ms window must be carefully inspected. With our model, inference takes 10 seconds per minute of recording, and expert verification for false positives adds roughly 30 seconds (~8 detected windows to check, given sensitivity 0.4 and specificity 0.99). This reduces the total burden to ~40 seconds per minute, which is six times faster than manual review. Our results show that training our models on the refined annotations, instead of the less-precise original ones, does not significantly improve overall classification performance. This suggests that high quality annotations may not be necessary to train reasonably performing models. Nevertheless, we observe an increased sensitivity suggesting that cleaner annotations allow the detection of additional spikes that may be less visible in the signal. We suspect that the extreme class imbalance in our data is the primary reason why our results do not improve significantly when trained using refined annotations. Future work could aim at weighting more heavily, during training, the samples whose annotations changed during the iterative learning procedure. Doing so would amplify their impact and potentially overcome the performance plateau as these samples may reflect greater information value.

Annotation quality: Our results revealed that the annotation quality was less critical for model training than we initially expected. Stable performance across the annotation refinement steps suggest that deep learning models can extract useful patterns even from imperfect or non-exhaustive labels. This counterintuitive finding was previously observed in a similar task of seizure detection in EEG by (Saab et al., 2020), where the authors observed that a large, poorly annotated dataset yielded better results than a smaller, high-quality annotated dataset. However, annotation quality remains essential for reliable evaluation. In addition, our results suggest that our original noisy annotations were not random, as we observe a sharper drop in performance in the random noise addition experiment. As a result, deep learning models could potentially benefit from being trained on a larger database with labels obtained from traditional baseline models.

Limitations: We followed a classification approach using a windowing strategy of the data to detect interictal spikes. This approach leads to a high data imbalance problem making model weight optimization more challenging. Alternative approaches based on anomaly or event detection could be more appropriate for this task (Luca et al., 2014; Zamanzadeh Darban et al., 2024), although they were not directly applicable to our original data as the annotations were incomplete. Future work should explore alternative model architectures better suited to the nature of the data, including transformers (Wen et al., 2023). In addition, our study was limited to the detection of interictal spikes, and we did not perform source localization to obtain the epileptogenic zone. Such end-to-end pipeline based on deep learning already shown promising results for MEG data (Hirano et al., 2022; Zheng et al., 2023) and represents a future research direction.

## 5. Conclusion

To conclude, we propose two machine learning models for accurate interictal spike detection from raw MEG data, alongside a strategy for refining electrophysiology data annotations. Our results suggest that, when working with heterogeneous realistic data, simpler models tend to perform best. Additionally, high annotation quality might not be necessary to train best performing machine learning models. However, clean annotations remain crucial for reliable model evaluation. Accurate localization of the epileptogenic zone, especially through non-invasive recordings, is essential for successful surgical treatment in pharmacoresistant epilepsy. Our approach, which automatically detects interictal spikes from routine clinical data, combined with source localization in future work, holds promise for improving this process. Our model architectures and interactive machine learning strategy could be readily extended to a variety of other applications based on EEG and MEG signals.

REFERENCES

Abd El-Samie, F. E., Alotaiby, T. N., Khalid, M. I., Alshebeili, S. A., & Aldosari, S. A. (2018). A Review of EEG and MEG Epileptic Spike Detection Algorithms. *IEEE Access*, *6*, 60673–60688. IEEE Access. https://doi.org/10.1109/ACCESS.2018.2875487

Alotaiby, T. N., Alrshoud, S. R., Alshebeili, S. A., Alhumaid, M. H., & Alsabhan, W. M. (2017). Epileptic MEG Spike Detection Using Statistical Features and Genetic Programming with KNN. *Journal of Healthcare Engineering*, *2017*, 3035606. https://doi.org/10.1155/2017/3035606

Amershi, S., Cakmak, M., Knox, W. B., & Kulesza, T. (2014). Power to the People: The Role of Humans in Interactive Machine Learning. *AI Magazine*, *35*(4), 105–120. https://doi.org/10.1609/aimag.v35i4.2513

Beghi, E. (2020). The Epidemiology of Epilepsy. *Neuroepidemiology*, *54*(2), 185–191. https://doi.org/10.1159/000503831

Bell, G. S., Neligan, A., & Sander, J. W. (2014). An unknown quantity—The worldwide prevalence of epilepsy. *Epilepsia*, *55*(7), 958–962. https://doi.org/10.1111/epi.12605

Budd, S., Robinson, E. C., & Kainz, B. (2021). A survey on active learning and human-in-the-loop deep learning for medical image analysis. *Medical Image Analysis*, *71*, 102062. https://doi.org/10.1016/j.media.2021.102062

da Silva Lourenço, C., Tjepkema-Cloostermans, M. C., & van Putten, M. J. A. M. (2021). Machine learning for detection of interictal epileptiform discharges. *Clinical Neurophysiology*, *132*(7), 1433–1443. https://doi.org/10.1016/j.clinph.2021.02.403

De Tiège, X., Lundqvist, D., Beniczky, S., Seri, S., & Paetau, R. (2017). Current clinical magnetoencephalography practice across Europe: Are we closer to use MEG as an established clinical tool? *Seizure - European Journal of Epilepsy*, *50*, 53–59. https://doi.org/10.1016/j.seizure.2017.06.002

Diachenko, M., Houtman, S. J., Juarez-Martinez, E. L., Ramautar, J. R., Weiler, R., Mansvelder, H. D., Bruining, H., Bloem, P., & Linkenkaer-Hansen, K. (2022). Improved Manual Annotation of EEG Signals through Convolutional Neural Network Guidance. *eNeuro*, *9*(5). https://doi.org/10.1523/ENEURO.0160-22.2022

El Ouahidi, Y., Gripon, V., Pasdeloup, B., Bouallegue, G., Farrugia, N., & Lioi, G. (2024). A Strong and Simple Deep Learning Baseline for BCI Motor Imagery Decoding. *IEEE Transactions on Neural Systems and Rehabilitation Engineering*, *32*, 3338–3347. IEEE Transactions on Neural Systems and Rehabilitation Engineering. https://doi.org/10.1109/TNSRE.2024.3451010

Fernández-Martín, R., Gijón, A., Feys, O., Juvené, E., Aeby, A., Urbain, C., Tiège, X. D., & Wens, V. (2024). *STIED: A deep learning model for the SpatioTemporal detection of focal Interictal Epileptiform Discharges with MEG* (arXiv:2410.23386). arXiv. https://doi.org/10.48550/arXiv.2410.23386

Gao, J., Song, X., Wen, Q., Wang, P., Sun, L., & Xu, H. (2021). *RobustTAD: Robust Time Series Anomaly Detection via Decomposition and Convolutional Neural Networks* (arXiv:2002.09545). arXiv. https://doi.org/10.48550/arXiv.2002.09545

Göndöcs, D., & Dörfler, V. (2024). AI in medical diagnosis: AI prediction & human judgment. *Artificial Intelligence in Medicine*, *149*, 102769. https://doi.org/10.1016/j.artmed.2024.102769

He, H., Liu, X., & Hao, Y. (2021). A progressive deep wavelet cascade classification model for epilepsy detection. *Artificial Intelligence in Medicine*, *118*, 102117. https://doi.org/10.1016/j.artmed.2021.102117

Hirano, R., Emura, T., Nakata, O., Nakashima, T., Asai, M., Kagitani-Shimono, K., Kishima, H., & Hirata, M. (2022). Fully-Automated Spike Detection and Dipole Analysis of Epileptic MEG Using Deep Learning. *IEEE Transactions on Medical Imaging*, *41*(10), 2879–2890. IEEE Transactions on Medical Imaging. https://doi.org/10.1109/TMI.2022.3173743

Jing, J., Herlopian, A., Karakis, I., Ng, M., Halford, J. J., Lam, A., Maus, D., Chan, F., Dolatshahi, M., Muniz, C. F., Chu, C., Sacca, V., Pathmanathan, J., Ge, W., Sun, H., Dauwels, J., Cole, A. J., Hoch, D. B., Cash, S. S., & Westover, M. B. (2020). Interrater Reliability of Experts in Identifying Interictal Epileptiform Discharges in Electroencephalograms. *JAMA Neurology*, *77*(1), 49–57. https://doi.org/10.1001/jamaneurol.2019.3531

Jung, J., Bouet, R., Delpuech, C., Ryvlin, P., Isnard, J., Guenot, M., Bertrand, O., Hammers, A., & Mauguière, F. (2013). The value of magnetoencephalography for seizure-onset zone localization in magnetic resonance imaging-negative partial epilepsy. *Brain*, *136*(10), 3176–3186. https://doi.org/10.1093/brain/awt213

Kljajic, J., O'Toole, J. M., Hogan, R., & Skoric, T. (2025). *Honest and Reliable Evaluation and Expert Equivalence Testing of Automated Neonatal Seizure Detection* (arXiv:2508.04899). arXiv. https://doi.org/10.48550/arXiv.2508.04899

Lin, T.-Y., Goyal, P., Girshick, R., He, K., & Dollar, P. (2017). Focal Loss for Dense Object Detection. *Proceedings of the IEEE International Conference on Computer Vision*, 2980–2988. https://openaccess.thecvf.com/content_iccv_2017/html/Lin_Focal_Loss_for_ICCV_2017_paper.html

Luca, S., Karsmakers, P., Cuppens, K., Croonenborghs, T., Van de Vel, A., Ceulemans, B., Lagae, L., Van Huffel, S., & Vanrumste, B. (2014). Detecting rare events using extreme value statistics applied to epileptic convulsions in children. *Artificial Intelligence in Medicine*, *60*(2), 89–96. https://doi.org/10.1016/j.artmed.2013.11.007

Mohammed, A. H., Cabrerizo, M., Pinzon, A., Yaylali, I., Jayakar, P., & Adjouadi, M. (2023). Graph neural networks in EEG spike detection. *Artificial Intelligence in Medicine*, *145*, 102663. https://doi.org/10.1016/j.artmed.2023.102663

Mosqueira-Rey, E., Hernández-Pereira, E., Alonso-Ríos, D., Bobes-Bascarán, J., & Fernández-Leal, Á. (2023). Human-in-the-loop machine learning: A state of the art. *Artificial Intelligence Review*, *56*(4), 3005–3054. https://doi.org/10.1007/s10462-022-10246-w

Mouches, P., Dejean, T., Jung, J., Bouet, R., Lartizien, C., & Quentin, R. (2024). Time CNN and Graph Convolution Network for Epileptic Spike Detection in Meg Data. *2024 IEEE International Symposium on Biomedical Imaging (ISBI)*, 1–5. https://doi.org/10.1109/ISBI56570.2024.10635822

Ossenblok, P., De Munck, J. C., Colon, A., Drolsbach, W., & Boon, P. (2007). Magnetoencephalography Is More Successful for Screening and Localizing Frontal Lobe Epilepsy than Electroencephalography. *Epilepsia*, *48*(11), 2139–2149. https://doi.org/10.1111/j.1528-1167.2007.01223.x

Peng, H., Gong, W., Beckmann, C. F., Vedaldi, A., & Smith, S. M. (2021). Accurate brain age prediction with lightweight deep neural networks. *Medical Image Analysis*, *68*, 101871. https://doi.org/10.1016/j.media.2020.101871

Raghu, M., Zhang, C., Kleinberg, J., & Bengio, S. (2019). Transfusion: Understanding Transfer Learning for Medical Imaging. *Advances in Neural Information Processing Systems (NeurIPS)*, *32*.

Rampp, S., Stefan, H., Wu, X., Kaltenhäuser, M., Maess, B., Schmitt, F. C., Wolters, C. H., Hamer, H., Kasper, B. S., Schwab, S., Doerfler, A., Blümcke, I., Rössler, K., & Buchfelder, M. (2019). Magnetoencephalography for epileptic focus localization in a series of 1000 cases. *Brain*, *142*(10), 3059–3071. https://doi.org/10.1093/brain/awz231

Roy, Y., Banville, H., Albuquerque, I., Gramfort, A., Falk, T. H., & Faubert, J. (2019). Deep learning-based electroencephalography analysis: A systematic review. *Journal of Neural Engineering*, *16*(5), 051001. https://doi.org/10.1088/1741-2552/ab260c

Ryvlin, P., Cross, J. H., & Rheims, S. (2014). Epilepsy surgery in children and adults. *The Lancet. Neurology*, *13*(11), 1114–1126. https://doi.org/10.1016/S1474-4422(14)70156-5

Saab, K., Dunnmon, J., Ré, C., Rubin, D., & Lee-Messer, C. (2020). Weak supervision as an efficient approach for automated seizure detection in electroencephalography. *Npj Digital Medicine*, *3*(1), 1–12. https://doi.org/10.1038/s41746-020-0264-0

Scott, J. M., Robinson, S. E., Holroyd, T., Coppola, R., Sato, S., & Inati, S. K. (2016). Localization of interictal epileptic spikes with MEG: Optimization of an automated beamformer screening method (SAMepi) in a diverse epilepsy population. *Journal of Clinical Neurophysiology : Official Publication of the American Electroencephalographic Society*, *33*(5), 414. https://doi.org/10.1097/WNP.0000000000000255

Smilkov, D., Thorat, N., Kim, B., Viégas, F., & Wattenberg, M. (2017). *SmoothGrad: Removing noise by adding noise* (arXiv:1706.03825). arXiv. https://doi.org/10.48550/arXiv.1706.03825

Tarassenko, L., Khan, Y. U., & Holt, M. R. G. (1998). Identification of inter-ictal spikes in the EEG using neural network analysis. *IEE Proceedings - Science, Measurement and Technology*, *145*(6), 270–278. https://doi.org/10.1049/ip-smt:19982328

Wen, Q., Zhou, T., Zhang, C., Chen, W., Ma, Z., Yan, J., & Sun, L. (2023). Transformers in time series: A survey. *Proceedings of the Thirty-Second International Joint Conference on Artificial Intelligence*, 6778–6786. https://doi.org/10.24963/ijcai.2023/759
16

Zamanzadeh Darban, Z., Webb, G. I., Pan, S., Aggarwal, C., & Salehi, M. (2024). Deep Learning for Time Series Anomaly Detection: A Survey. *ACM Comput. Surv.* https://doi.org/10.1145/3691338

Zheng, L., Liao, P., Luo, S., Sheng, J., Teng, P., Luan, G., & Gao, J.-H. (2020). EMS-Net: A Deep Learning Method for Autodetecting Epileptic Magnetoencephalography Spikes. *IEEE Transactions on Medical Imaging*, *39*(6), 1833–1844. IEEE Transactions on Medical Imaging. https://doi.org/10.1109/TMI.2019.2958699

Zheng, L., Liao, P., Wu, X., Cao, M., Cui, W., Lu, L., Xu, H., Zhu, L., Lyu, B., Wang, X., Teng, P., Wang, J., Vogrin, S., Plummer, C., Luan, G., & Gao, J.-H. (2023). An artificial intelligence–based pipeline for automated detection and localisation of epileptic sources from magnetoencephalography. *Journal of Neural Engineering*, *20*(4), 046036. https://doi.org/10.1088/1741-2552/acef92

STATEMENTS AND DECLARATIONS

**Funding**

This work was supported by a postdoctoral fellowship from the Fondation pour la Recherche Médicale, grant number SPF202209015835, to Pauline Mouches.

**Competing interests**

The authors declare that they have no known competing financial interests or personal relationships that could have appeared to influence the work reported in this paper.

**Author contributions**

**P.M.**: Conceptualization, Funding acquisition, Methodology, Formal analysis, Writing – original draft, **J.J.**: Conceptualization, Data curation, Funding acquisition, Validation, Writing – review and editing, **A.D.**: Methodology, Formal analysis, **A.G.**: Methodology, Writing – review and editing, **R.B.:** Conceptualization, Data curation, Validation, **R.M**: Data curation, Validation, **R.Q.**: Conceptualization, Funding acquisition, Supervision, Writing – review and editing.

**Data and Code Availability**

The code of the proposed models will be made available upon acceptance of the manuscript, along with example data. Source MEG data used in this study cannot be made publicly available.

**Ethics approval**

This study was performed in line with the principles of the Declaration of Helsinki. Approval was granted by the Ethics Committee of the Hospices Civils de Lyon on October 2012 (study NCT01735032: Multimodal Imaging in Pre-surgical Evaluation of Epilepsy (EPIMAGE)).

**Consent to participate**

Informed consent was obtained from all individual participants included in the study.